\documentclass[conference]{IEEEtran}
\IEEEoverridecommandlockouts
\usepackage{cite}
\usepackage{amsmath,amssymb,amsfonts}
\usepackage{algorithmic}
\usepackage{graphicx}
\usepackage{textcomp}
\usepackage{xcolor}
\usepackage{balance}
\usepackage{subcaption}
\usepackage[symbol]{footmisc}
\def\BibTeX{{\rm B\kern-.05em{\sc i\kern-.025em b}\kern-.08em
    T\kern-.1667em\lower.7ex\hbox{E}\kern-.125emX}}
\begin{document}

\title{Towards an Integrated Performance Framework for Fire Science and Management Workflows
}

\author{\IEEEauthorblockN{Hena Ahmed\IEEEauthorrefmark{1}}
\IEEEauthorblockA{\textit{Hal{\i}c{\i}o\u{g}lu Data Science Institute} \\
\textit{University of California, San Diego}\\
La Jolla, CA, USA \\
h7ahmed@ucsd.edu}
\and
\IEEEauthorblockN{Ravi Shende\IEEEauthorrefmark{1}}
\IEEEauthorblockA{\textit{San Diego Supercomputer Center} \\
\textit{University of California, San Diego}\\
La Jolla, CA, USA \\
rshende@ucsd.edu}
\and
\IEEEauthorblockN{Ismael Perez}
\IEEEauthorblockA{\textit{San Diego Supercomputer Center} \\
\textit{University of California, San Diego}\\
La Jolla, CA, USA \\
i3perez@sdsc.edu}
\and
\IEEEauthorblockN{Daniel Crawl}
\IEEEauthorblockA{\textit{San Diego Supercomputer Center} \\
\textit{University of California, San Diego}\\
La Jolla, CA, USA \\
lcrawl@ucsd.edu}
\and
\IEEEauthorblockN{Shweta Purawat}
\IEEEauthorblockA{\textit{San Diego Supercomputer Center} \\
\textit{University of California, San Diego}\\
La Jolla, CA, USA \\
shpurawat@ucsd.edu}
\and
\IEEEauthorblockN{\.{I}lkay Altinta\c{s}}
\IEEEauthorblockA{\textit{San Diego Supercomputer Center and} \\
\textit{Hal{\i}c{\i}o\u{g}lu Data Science Institute} \\
\textit{University of California, San Diego}\\
La Jolla, CA, USA \\
ialtintas@ucsd.edu}
}

\maketitle
\begingroup\renewcommand\thefootnote{\IEEEauthorrefmark{1}}
\footnotetext{Equal contribution}
\endgroup

\begin{abstract}
Reliable performance metrics are necessary prerequisites to building large-scale end-to-end integrated workflows for collaborative scientific research, particularly within context of use-inspired decision making platforms with many concurrent users and when computing real-time and urgent results using large data. This work is a building block for the National Data Platform, which leverages multiple use-cases including the WIFIRE Data and Model Commons for wildfire behavior modeling and the EarthScope Consortium for collaborative geophysical research. 
This paper presents an artificial intelligence and machine learning (AI/ML) approach to performance assessment and optimization of scientific workflows. 
An associated early AI/ML framework spanning performance data collection, prediction and optimization is applied to  wildfire science applications within the WIFIRE BurnPro3D (BP3D) platform for proactive fire management and mitigation. 
\end{abstract}



\section{Introduction}
Scientific application workflows have become a key tool in natural disaster mitigation and response. Real-time sensor and satellite data now provide invaluable resources for urgent science analytics to be conducted with remarkable speed and precision. Workflows and larger cyberinfrastructures (CIs) powered by such data can deliver critical knowledge about imminent natural hazards such as wildfires~\cite{gollner2015towards}, earthquakes~\cite{dittmann2022comparing}, and volcanic eruptions~\cite{corsa2022integration}. 

However, the vast influx of raw and pre-processed data from geo-distributed sources presents challenges to the design of scalable cyberinfrastructures for data to knowledge workflows, thus heightening the need for developing a computing continuum of integrated cloud-to-edge resources~\cite{balouek2019towards}. A computing continuum especially enables novel implementations of urgent application workflow with particular attention to efficient data processing, and reliable but timely data to knowledge transfer to support urgent decision-making~\cite{balouek2020harnessing}.

\begin{figure}[htbp]
    \centering
    \includegraphics[width=\linewidth]{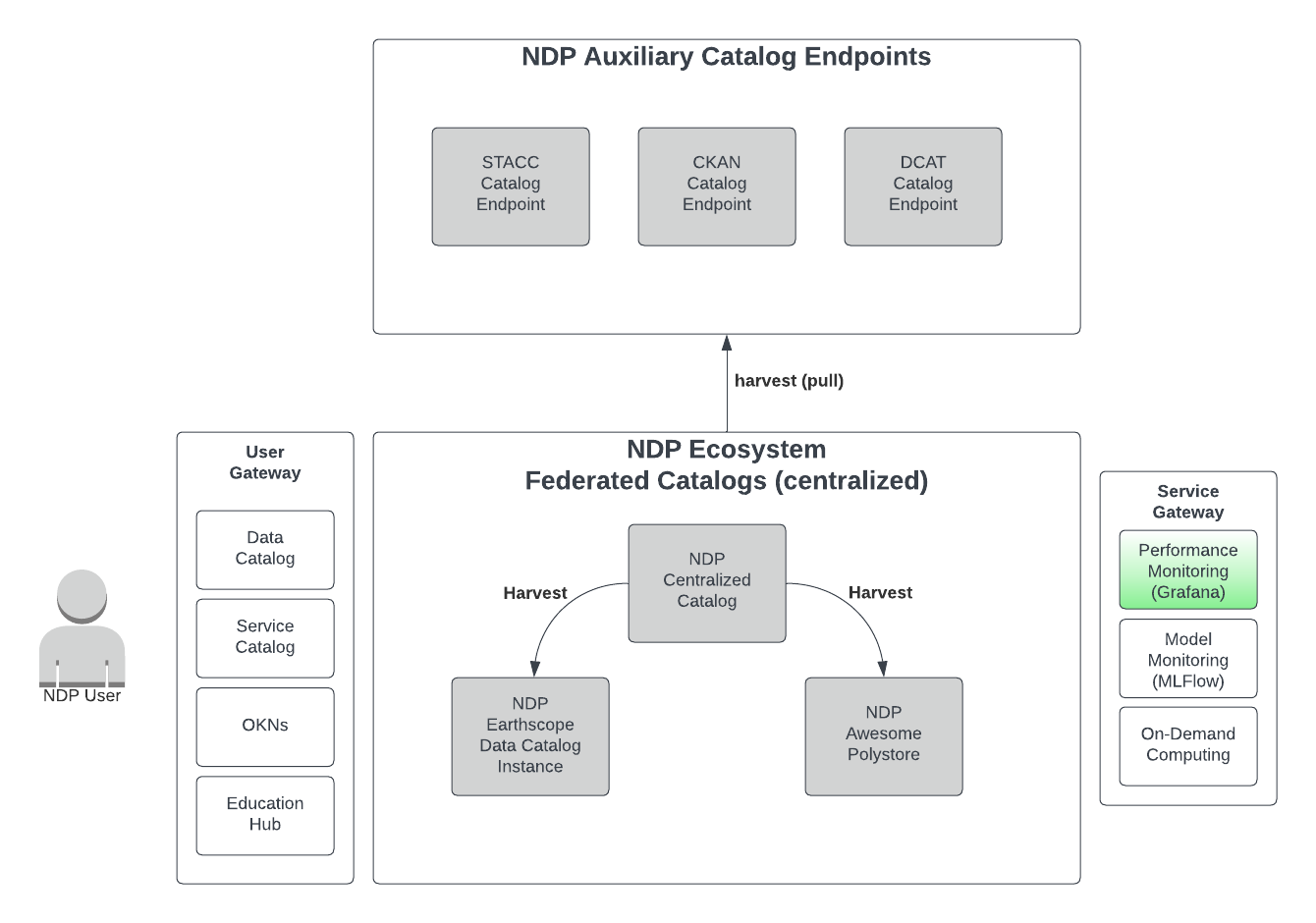}
    \caption{}
    \label{fig:ndp}
\end{figure}

Earlier works presented the WIFIRE cyberinfrastructure of integrated end-to-end workflows for wildfire behavior modeling~\cite{altintas2015towards}, as well as a use case of the computing continuum to support data-driven workflows for air quality prediction to manage wildfire impacts~\cite{balouek2023keynote}. The WIFIRE Commons itself is one such use-case for the National Data Platform project, which will leverage the computing continuum to democratize scientific data access and analysis through a national cyberinfrastructure~\cite{parashar2019virtual}. Fig~\ref{fig:ndp} outlines an early performance pipeline for the National Data Platform (NDP).

Our work particularly addresses the issue of performance optimization for the WIFIRE-powered BurnPro3D (BP3D) platform~\cite{bp3d, roten2023truetrees}. BP3D is a decision support platform to inform and optimize prescribed burn planning for wildfire management. The platform works in tandem with other WIFIRE frameworks, namely QUIC-Fire fire and atmospheric models and FastFuels 3D fuel structure models, in order to identify environmental conditions and ignition patterns that are optimal for prescribed burns. 

BP3D is a user-facing tool for geographically distributed land managers and fire planners, and is applied in a variety of computing environments with different capabilities. In this paper, we present preliminary steps towards a performance prediction framework that will be used by BP3D users to assess the necessary resource provisions to run BP3D given environmental and fuel data inputs. We present a performance framework for integrating data processing and AI/ML techniques in order to predict resource consumption during BP3D runs, and thereby improve the scalability and reliability of underlying BP3D data workflows. We implement a data processing architecture to collect and prepare performance data for AI/ML analysis, and demonstrate examples of predictive modeling techniques for performance evaluation and provision of our AI-driven data workflows. This is part of a larger work to optimize sub-systems for integrated end-to-end data cyberinfrastructures, which is critical to enabling efficient data processing and modeling across the computing continuum.

The rest of the paper is structured as follows: Section~\ref{sec:related-works} discusses related works in the existing scientific literature, Section~\ref{sec:use-case} describes a use case application of our workflow architecture, Section~\ref{sec:approach} describes our methodological approach and design, Section~\ref{sec:results} demonstrates early results of our framework, and Section~\ref{sec:conclusion} summarizes conclusions and future endeavors of this work. 
\section{Related Works}
\label{sec:related-works}

Large-scale data cyberinfrastructures are next generation platforms for collaborative research workflow and data sharing. Beckman et al.~\cite{doi:https://doi.org/10.1002/9781119551713.ch7} highlights the need for performance optimization across the computing continuum. Current case studies for scalable CIs include the Virtual Data Collaboratory for interdisciplinary data and science sharing presented by Parashar et al.~\cite{parashar2019virtual}, and the EarthScope~\cite{corsa2022integration} framework for open access, real-time geophysical data, modeling, and educational services. Our work also draws from contributions in distributed and multimodal data architectures such as the Quantum Data Hub presented by Purawat et al.~\cite{purawat2021quantum} and the AWESOME polystore using open-knowledge networks (OKNs) presented by Dasgupta \& Gupta~\cite{dasgupta2016analytics}.

 Nguyen et al.~\cite{nguyen2016integrated} presented methods for integrating machine learning techniques in scientific workflow systems to evaluate accuracy and scalability. As described in Parashar et al.~\cite{parashar2019virtual}, the GeoSciFramework (commonly known as EarthScope) is one such case study for scalable architectures of scientific workflows and integrative machine learning environments that operate with continuously streaming geodetic and seismic data. However, an existing key problem area in developing scalable architectures for integrated machine learning and scientific workflows is developing knowledge management techniques to assimilate and prepare data from sources for AI/ML analysis. This concept is also referred to as the ``AI-readiness'' of data. AI-readiness is especially important for urgent computing applications such as natural hazard modeling and prediction-making. Baru et al.~\cite{baru2022enabling} are currently addressing the challenge of finding and matching AI-ready data and models in an integrated platform, while also following guidelines for FAIR~\cite{wilkinson2016fair} for data provenance. Holding scientific work to FAIR data management principles~ (where FAIR stands for Findability, Accessibility, Interoperability, and Reusability) is a key step to ensuring the responsible deployment of AI models and other data services.

By building upon existing work in integrated AI/ML and scientific workflow architectures as well, this paper will further previous research towards global, integrated cyberinfrastructures that enable equitable data-driven technology sharing.
\section{Workflow Use Case}\label{sec:use-case}

The use case application described in this section was created to predict total resource consumption of BurnPro3D simulations for prescribed burns and wildfire mitigation~\cite{bp3d}. An execution of BP3D takes a single set of environmental input data and runs an ensemble of simulations over multiple Kubernetes servers. We created an integrated ML/AI workflow that retrieves the input parameter values given to a BP3D run and resource consumption data that is generated throughout the run and stored on Nautilus servers. The workflow then takes a linear regression approach to predicting total CPU and memory usage of a BP3D run. 

For the purposes of this paper, the chosen machine learning method (linear regression) is rather arbitrary, as our intentions at this stage of research are to demonstrate a functional data-driven pipeline for AI/ML performance analysis, as opposed to choosing the most accurate or robust modeling approach for the task. So, we used a basic linear regression implementation with default model parameter values made accessible using the popular Python package Scikit-Learn. In future extensions of this work, we plan to conduct more extensive sensitivity analyses and parameter estimation techniques in order to implement modeling techniques that better fit the given data. The results of this particular use case are discussed in Section~\ref{sec:results}. 
 
\section{Approach}  
\label{sec:approach}

In this section, we describe our methodology for retrieving resource consumption data and integrating AI/ML-informed decision-making to analyze overall performance of the BurnPro3D architecture for prescribed burn modeling. We describe two primary objectives in developing our AI/ML solution: 1) data preparation for AI-readiness (Section~\ref{subsec:approach-ai-readiness}), and 2) integration of predictive ML/AI modeling methods into the scientific workflow architecture of BP3D (Section~\ref{subsec:approach-model-design}). 

    \subsection{AI-Ready Data Preparation}\label{subsec:approach-ai-readiness}
    We first discuss the steps that were taken towards reaching our AI-readiness objective. Data assimilation and preparation requirements 
    will vary depending on the form of ML/AI analysis being applied. The use case we describe in this paper demonstrates results obtained through linear regression. So, in this section, we discuss the steps taken to achieve data that is AI-ready in the specific context of linear regression modeling. 
    
    By accessing the BurnPro3D API, we can retrieve data about individual ensembles of simulation runs generated using the QUIC-FIRE model. This paper deals with two different classes of data relevant to BurnPro3D simulations: first are the model input data features (weather and atmospheric data, ignition and fuel conditions, geospatial data) for BP3D models; second are performance data (runtime, CPU usage, memory usage, storage I/O, network usage) generated during a BurnPro3D ensemble run.

    Each BP3D ensemble is hosted on one or more Kubernetes nodes, where simulation runs are hosted on different pods in the node. The performance data is then stored as time series data on a Nautilus server, where it can be queried using PromQL (a querying language for Prometheus servers in Nautilus) and visualized on a Prometheus web user interface and/or Grafana dashboard.

    Prior to retrieving performance data, we store the following identifiers for each BP3D run: the set of all input parameters passed into the simulation, the unique and corresponding ensemble IDs (i.e., the pod/node pair in Kubernetes), and total simulation runtime. Once these identifiers are stored, we proceed to collect the start/end timestamps and input parameter values for each run per ensemble. This data set will later be used as training data for performance optimizing AI/ML methods. Occasionally, a simulation run will fail, as indicated in the data by \texttt{NA} timestamps in the start, end, or total simulation time categories. We chose not to include failed runs in the final training data set. Table~\ref{tab:bp3d-inputs} describes input and output features collected after a simulation run of BurnPro3D, and Table~\ref{tab:performance-data} describes performance features retrieved during the run.

    \begin{table}
      \centering
      \caption{BurnPro3D Inputs/Outputs}
      \begin{tabular}{|cl|}
        \hline
        \textbf{Feature Name} & \textbf{Description} \\
        \hline
        \texttt{surface\_moisture} & surface fuel moisture \\
        \texttt{wind\_moisture} & fuel moisture of surface winds \\
        \texttt{wind\_direction} & direction of surface winds \\
        \texttt{wind\_speed} & speed of surface winds \\
        \texttt{sim\_time} & estimated minimum runtime (seconds)\\
        \texttt{timestep} & elapsed seconds between simulation steps \\
        \texttt{run\_max\_mem\_rss\_bytes} & maximum RSS bytes allowed per run \\
        \texttt{area} & calculated regional surface area\\
        \texttt{runtime} & time for whole run simulation (seconds)\\
        \hline
      \end{tabular}
      \label{tab:bp3d-inputs}
    \end{table}

    \begin{table}
      \centering
      \caption{Performance Outputs}
      \begin{tabular}{|cl|}
        \hline
        \textbf{Feature Name} & \textbf{Description} \\
        \texttt{pod} & unique ID of a Kubernetes pod \\
        \texttt{node} & unique ID of a Kubernetes node \\
        \texttt{start} & datetime-stamp marking the beginning of a run \\
        \texttt{stop} & datetime-stamp marking the end of a run \\
        \texttt{threads} & total \# of threads used \\
        \texttt{memory\_requests} & Min bytes of memory requested\\
        \texttt{cpu\_usage} & total CPU usage seconds \\
        \texttt{mem\_usage} & Max bytes of memory used \\
        \hline
      \end{tabular}
      \label{tab:performance-data}
    \end{table}

    We can then retrieve the performance data that was collected and stored in our Nautilus server during each simulation run, and pre-process the JSON-formatted data to achieve tabular data sets suitable for basic AI/ML models. For our experiments, we are focusing on resource consumption over the duration of a BP3D simulation run. So, we query for the minimum memory requested for each pod, the total CPU usage, and total memory usage during the time range of each run. Total CPU usage and memory usage of a BP3D simulation will be the target variables that we want to predict using AI/ML modeling methods. We also collect the partial CPU and memory usage of a run from the start time until certain time periods (either predetermined or pseudo-random) to represent refreshed prediction times. This data will then be used with an AI/ML approach to predict the performance consumption metrics of the entire duration of each run.

    \subsection{Predictive Model Design}\label{subsec:approach-model-design}
    Having assembled an AI-ready data set, we now want implement a predictive model to evaluate the performance of BP3D simulations. This stage follows a standard experimental procedure of: 1) choosing a predictive modeling method, 2) data pre-processing and feature analysis, and 3) model fitting, training, and testing. 
    
\begin{figure}[htbp!]
    \centerline{\includegraphics[width=\linewidth]{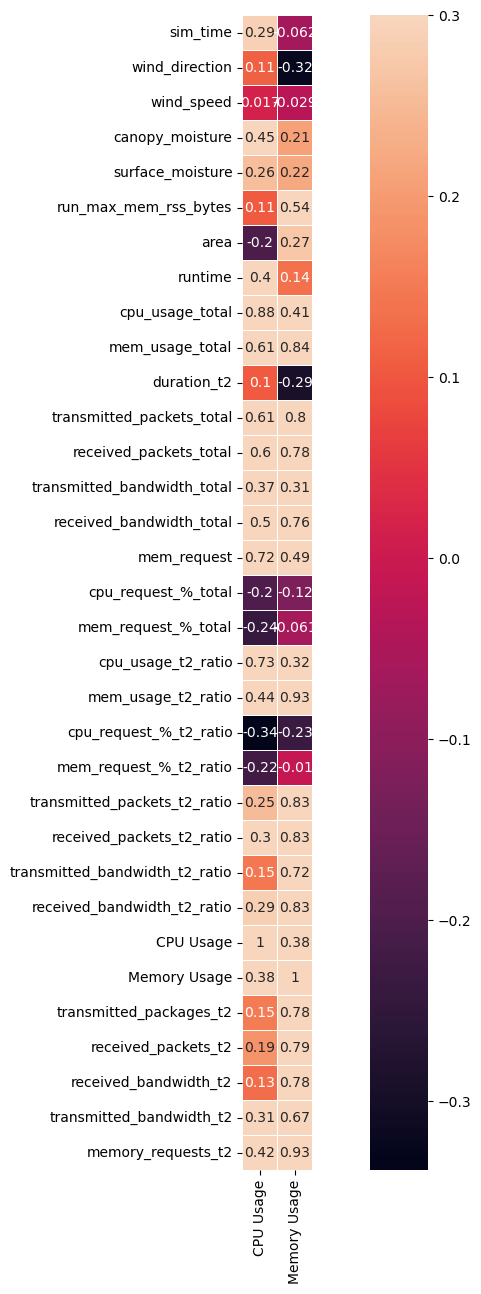}}
    \caption{Pearson correlation coefficients for CPU and Memory Usage.}
    \label{fig:corr-matrix}
    \end{figure}

    We have two target features in our data to indicate resource utilization during a given simulation run: total CPU usage (as a percentage of total CPU capacity) and total memory usage (as a percentage of total memory available). Before training the model, we calculated Pearson correlation coefficients to determine which of the variables (which consist of BP3D simulation input parameters and resource consumption data collected from a BP3D run) in our data set were most strongly indicative of changes in resource utilization. Figure~\ref{fig:corr-matrix} shows the correlation matrix for our assembled data set.

   We select features that are strongly correlated (i.e., with a Pearson coefficient $>.5$) with CPU Usage and/or Memory Usage. We then proceed to the model fitting stage, which would resemble the linear regression use case described in Section~\ref{sec:use-case}.

\subsection{User Experience Design}
    We consider two potential implementations for determining how users can interact with our predictive modeling workflow. One implementation would automatically refresh model predictions at set time intervals. To do this, we would automate queries to retrieve performance data of the current simulation over set time ranges (e.g. from the start of the run until 5 minutes in), and then rerunning the predictive model, taking into account the most recent query results in order to obtain an updated prediction of resource utilization over the entire simulation run.

    A second potential implementation would enable users to refresh performance predictions at any time point after the model has started running, and then continue to update the prediction as desired. There is a brief time period where the run has started, but the user is not able to request a prediction, as it takes time for data to be put onto Kubernetes servers. Our approach is to wait 45 seconds after the run starts, then automatically refresh the prediction. From then onwards, we allow the user to refresh the prediction as desired. This design approach would be similar to the previously described implementation, except that updated performance data would only be queried when/if the user indicates, rather than at set time intervals. Using this approach, the data collected at each model refresh would consist of performance metrics from the start time of the simulation to the refresh time. The data collected from the start until the refresh time would be used to obtain an updated prediction of resource utilization over the entire simulation run. We simulate these refresh times in our training data by inserting duration columns (like duration\_t2 from Figure 2) and querying from the start time until the duration time, generating data such as transmitted\_packages\_t2. We feed these into the model to predict the final resource consumption metrics.

    The decision ultimately centers on balancing user experience and the effectiveness of model training. In the first implementation, the model is trained at consistent time intervals, which could improve its ability to recognize patterns in BP3D resource consumption, and as a result, enhance prediction accuracy. However, this implementation might negatively impact the user experience due to limited user control and potential frustration arising from the passive nature of updates. Conversely, the second approach offers the user greater flexibility, allowing them to update predictions as needed and providing a more interactive experience. While this approach might introduce complexities in model performance, we could potentially mitigate these with issues with strategic data manipulation—such as using a ratio of refresh time to resource metrics instead of analyzing these metrics separately. This ratio is shown in  Figure~\ref{fig:corr-matrix} with the \_t1\_ratio and \_t2\_ratio metrics. A hybrid approach, integrating the consistent training intervals of the first method and the user-initiated updates of the second could also be considered. This would provide a more balanced approach, though its increased complexity could result in a less intuitive user experience.

\section{Early Results}
\label{sec:results}

In the initial stages of the workflow, we retrieve resource consumption data from over 900 BP3D runs. The retrieved data will include performance statistics under four distinct categories: CPU usage, memory usage, network usage, and storage I/O. Post-processing, the tabular data will describe performance features for each BP3D run. In addition to the unique pod/node IDs for each simulation run, the final, cleaned performance database will contain the features listed in Tables~\ref{tab:cpu-data}~-~\ref{tab:storage-data}.

\begin{table}[b]
      \centering
      \caption{CPU Quota}
      \begin{tabular}{|cl|}
        \hline
        \textbf{Feature Name} & \textbf{Description} \\
        \hline
        \texttt{CPU Usage} & total CPU usage seconds \\
        \texttt{CPU Requests} & number of CPU cores requested  \\
        \texttt{CPU Requests \%} & (CPU user time) / (total CPU time requested) \\
        \texttt{CPU Limits} & maximum capacity for CPU usage \\
        \texttt{CPU Limits \%} & (CPU user time) / (total CPU time limit) \\
        \hline
      \end{tabular}
      \label{tab:cpu-data}
    \end{table}

\begin{table}[b]
      \centering
      \caption{Memory Quota}
      \begin{tabular}{|cl|}
        \hline
        \textbf{Feature Name} & \textbf{Description} \\
        \hline
        \texttt{Memory Usage} & number of bytes of memory usage  \\
        \texttt{Memory Requests} & minimum number of bytes requested \\
        \texttt{Memory Requests \%} & (memory usage) / (memory requests) \\
        \texttt{Memory Limits} & maximum capacity for memory usage \\
        \texttt{Memory Limits \%} & (memory usage) / (memory limits) \\
        \texttt{Memory Usage (RSS)} & RSS bytes used  \\
        \texttt{Memory Usage (Cache)} & CPU cache memory used\\
        \hline
      \end{tabular}
      \label{tab:memory-data}
    \end{table}

\begin{table}[t]  
      \centering
      \caption{Network Usage}
      \hskip-1.4cm\begin{tabular}{|cl|}
        \hline
        \textbf{Feature Name} & \textbf{Description} \\
        \hline
        \texttt{Receive Bandwidth} &  network bandwidth for receiving bytes \\
        \texttt{Transmit Bandwidth} & network bandwidth for transmitting bytes \\
        \texttt{Received Packets} & \# of packets received in a run \\
        \texttt{Transmitted Packets} & \# of packets transmitted in a run \\
        \texttt{Received Packets Dropped} &  \# of received packets dropped in a run \\
        \texttt{Transmitted Packets Dropped} & \# of transmitted packets dropped in a run\\
        \hline
      \end{tabular}
      \label{tab:network-data}
    \end{table}

\begin{table}[t]
      \centering
      \caption{Current Storage IO}
      \hskip-1cm\begin{tabular}{|cl|}
        \hline
        \textbf{Feature Name} & \textbf{Description} \\
        \hline
        \texttt{IO (Reads)} & \# of I/Os read in a run  \\
        \texttt{IO (Writes)} & \# of I/Os written in a run \\
        \texttt{IO (Reads+Writes)} & \# of I/Os read and written in a run \\
        \texttt{Throughput (Read)} & bytes of throughput read during the run \\
        \texttt{Throughput (Write)} & bytes of throughput written in the run \\
        \texttt{Throughput (Read+Write)} & bytes of throughput read/written in a run \\
        \hline
      \end{tabular}
      \label{tab:storage-data}
    \end{table}

\begin{figure}[htb]
    \centering
\begin{subfigure}[htbp]{\textwidth}
    \includegraphics[width=.5\linewidth]{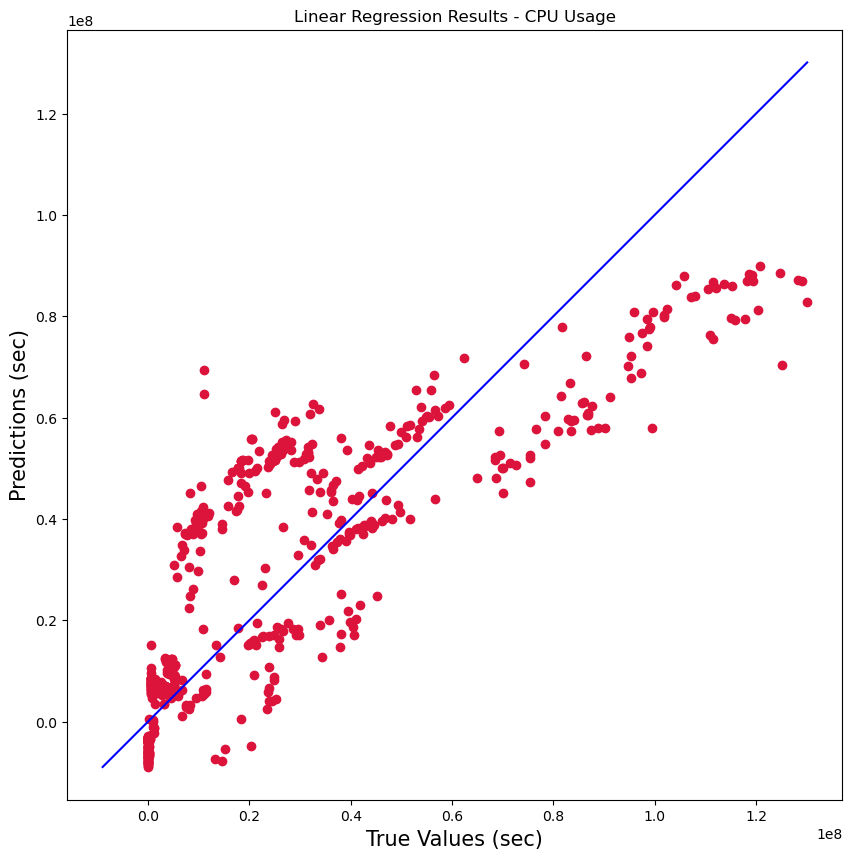}
    \label{fig:linreg-cpu}
    \end{subfigure}

\begin{subfigure}[htbp]{\textwidth}
    \includegraphics[width=.5\linewidth]{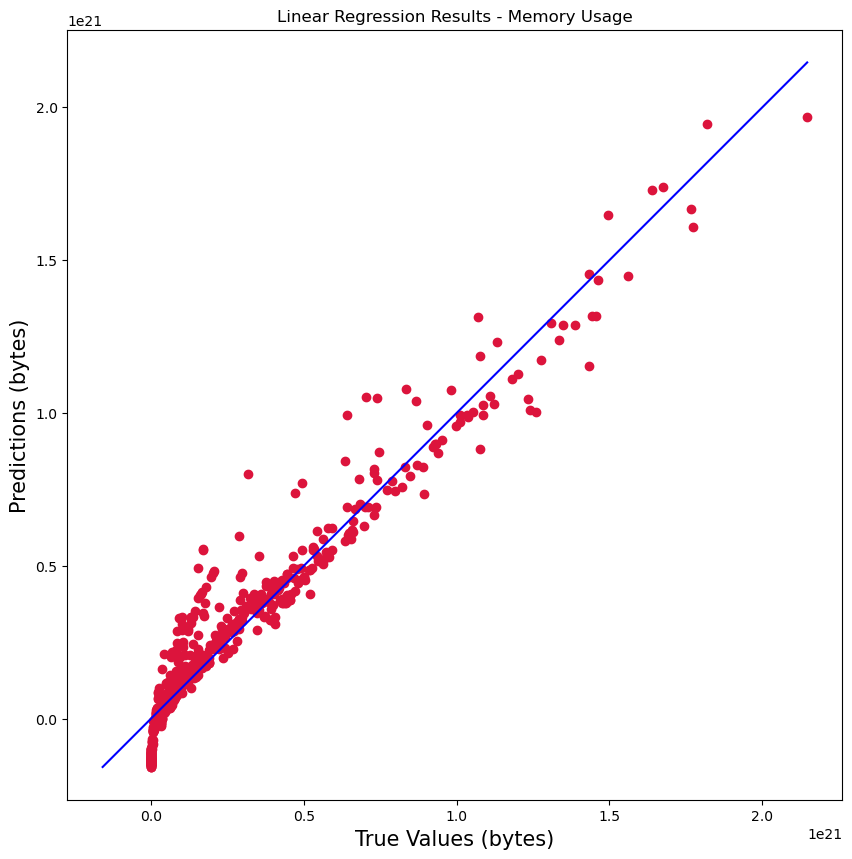}
    \label{fig:linreg-mem}
    \end{subfigure}

\caption{Preliminary results of linear regression use case (Sec~\ref{sec:use-case}).}
\label{fig:usecase-results}
    
\end{figure}

The next stage of the workflow (AI/ML modeling), outputs predictions of total CPU usage (in seconds) and memory usage (in bytes) during a BurnPro3D run. To demonstrate possible results at this stage, we return to the use case application of linear regression introduced in Section~\ref{sec:use-case}. For our linear regression model, we selected training features based on the feature analysis results described in Section~\ref{subsec:approach-model-design}. Figure~\ref{fig:usecase-results} shows the preliminary results of a linear regression model using sample performance data retrieved from BurnPro3D runs.

The linear regression model predicted CPU and memory usage with R-squared error rates of 0.70626 and 0.9221, respectively. However, given the shortage of training and testing data at this stage of our research, there are too few data points to obtain accurate or generalizable predictions through linear regression, and are currently working towards generating and preparing sufficient amounts of BP3D data in order to develop more robust AI/ML models. For the purposes of this paper, our demonstrated AI-readiness and AI/ML modeling pipelines are fundamental building blocks of an early workflow architecture.
\section{Conclusion \& Future Works}
\label{sec:conclusion}
We have presented our approach and early results of an integrated AI/ML workflow for performance analysis of the BurnPro3D fire management platform. Our use-case can be applied for the integration of AI-ready data preparation and AI/ML predictive modeling techniques in an end-to-end scientific workflow. where the use-case presented in this paper limits the use of ML/AI to identify relationships between BP3D input parameters and total resource consumption, our ongoing work aims to also optimize resource consumption for the purpose of mitigating uncertainty and improving accuracy of BP3D outputs. This work is part of a broader effort towards integrating AI/ML-driven methods for performance optimization in large cyberinfrastructures, namely the in-progress National Data Platform project. Immediate extensions of this work include the incorporation of a runtime prediction modeling stage into the current workflow, and the introduction of end-to-end uncertainty quantification metrics in order to align our work with FAIR data management standards for scientific research.

\section*{Acknowledgments} 
The authors would like to thank the WIFIRE and WorDS
teams for their collaboration and support of this study by NSF 2040676, 2134904 and 2333609, and  the Nautilus Kubernetes Cluster of the National Research Platform funded by in part by NSF awards 1730158, 1540112, 1541349, 1826967, 2112167 and 2120019.

\balance
\bibliographystyle{IEEEtran}
\bibliography{references}

\end{document}